# Image Compression with SVD : A New Quality Metric Based On Energy Ratio

[1] Henri Bruno Razafindradina, [2] Paul Auguste Randriamitantsoa, [3] Nicolas Raft Razafindrakoto

[1] Higher Institute of Technology
Diego-Suarez, 201, Madagascar

[2] Higher Polytechnic School of Antananarivo
Antananarivo, 101, Madagascar

[3] Higher Polytechnic School of Antananarivo
Antananarivo, 101, Madagascar

**Abstract** - Digital image compression is a technique that allows to reduce the size of an image in order to increase the capacity storage devices and to optimize the use of network bandwidth. The quality of compressed images with the techniques based on the discrete cosine transform or the wavelet transform is generally measured with PSNR or SSIM. Theses metrics are not suitable to images compressed with the singular values decomposition. This paper presents a new metric based on the energy ratio to measure the quality of the images coded with the SVD. A series of tests on 512 × 512 pixels images show that, for a rank k = 40 corresponding to a SSIM = 0,94 or PSNR = 35 dB, 99,9% of the energy are restored. Three areas of image quality assessments were identified. This new metric is also very accurate and could overcome the weaknesses of PSNR and SSIM.

**Keywords** - *Metric, Assessment, SVD, Singular Value, Image, Compression, PSNR, SSIM.*

## 1. Introduction

The Digital images compression knew a ceaseless evolution, since the 60s, in parallel with telecommunications and multimedia. It allows reducing the size of an image with the aim to increase the capacity storage of media (limited in capacity) and to optimize the use of the network bandwidth. Since the standardization of the JPEG algorithm based on the discrete cosine transform [1], the volume of multimedia data (sound, image, video, etc.) has steadily increased. The JPEG2000 standard based on the wavelet transform [2] increased compression ratio with a higher quality than JPEG. This article describes the digital images compression by compacting energy in the singular values domain. SVD consists on decomposing a matrix into a product of 3 matrices U, S and V (S is called the singular values matrix). Chen [3], Abrahamsen [4], and Kahu [5] have already proposed a gray level image compression method that retains only the first k singular values. The assessment of their method was only based on the measurement of the quadratic errors MSE or PSNR. Some methods used SSIM instead of PSNR.

We propose in this paper an assessment method based on a new metric called Energy Ratio. This metric computes the ratio between the energy of the image after compacting the singular values matrix and the energy of the original image. The basis of the coding and decoding technique will be first described, we will then detail the proposed metric and the results will be discussed.

## 2. Singular Values Decomposition

### 2.1 Theory

Any matrix I of size m × n of rank r can be decomposed into a weighted sum of unitary matrices m × n by Singular Value Decomposition (SVD). Matrices U and V are unitary and I can therefore be written :

$$I = U \times S \times V^T = \sum_{i=1}^{n} \left( \sigma_i \times u_i \times v_i^T \right) \qquad (1)$$

S is a matrix whose r first diagonal terms are positive, all the others were zero. The r non-zero terms $\sigma_i$ are called singular values of I.

$$S = \begin{pmatrix} \sigma_i & \cdots & 0 \\ \vdots & \ddots & \vdots \\ 0 & \cdots & \sigma_n \end{pmatrix} \quad (2)$$

Where $\sigma_1 \geq \sigma_2 \geq \ldots \geq \sigma_r$ et $\sigma_{r+1} \geq \sigma_{r+2} \geq \ldots \geq \sigma_n = 0$

2.2 SVD and Image Compression

The singular values represent the energy of the image. Sadek [6] speaks of SVD decomposition as an energy-oriented approach. Indeed, the total energy of the image I is represented in the following formula :

$$\|I\| = trace[I^T \times I] = \sum_{i=1}^{m}\sum_{j=1}^{n} I^2(i,j) = \sum_{i=1}^{n} \sigma_i^2 \quad (3)$$

The compression of a gray-level image thus comes intuitively by forcing the weakest singular values to zero. By selecting only the first singular values k (k ≤ n), we can approximate the matrix I by the formula :

$$I_k = U \times S_k \times V^T = \sum_{i=1}^{k} \sigma_i \times u_i \times v_i^T \quad (4)$$

$S_k$ represents the matrix of k compressed singular values. The corresponding energy is :

$$\|I_k\| = trace[I_k^T \times I_k] = \sum_{i=1}^{k} \sigma_i^2 \quad (5)$$

2.3 Assessment

PSNR: the PSNR (Peak Signal to Noise Ratio) has been used to measure the distortion between the compressed image and the original image. Let I be the original image and $I_k$ the compressed image.

$$PSNR = 10 \times \log_{10}\left(\frac{Max[I(i,j)]^2}{MSE}\right) \quad (6)$$

MSE is the Mean Squared Error :

$$MSE = \frac{1}{m \times n}\sum_{i=1}^{m}\sum_{j=1}^{n}[I(i,j) - I_k(i,j)]^2 \quad (7)$$

But since a few years, it has been replaced by SSIM, which better reproduces Human Visual Perception.

SSIM: The idea of SSIM [7] (Structural Similarity) is to compute the similarity of structure between both images, rather than a pixel-by-pixel difference as does the PSNR.

$$SSIM(I,J) = \frac{(2\mu_I\mu_J + C_1)(2\sigma_I\sigma_J + C_2)(2\,cov_{IJ} + C_3)}{(\mu_I^2 + \mu_J^2 + C_1)(\sigma_I^2 + \sigma_J^2 + C_2)(\sigma_I\sigma_J + C_3)} \quad (8)$$

- I the original image, J the compressed image ;
- $\mu_I$ the average of I, $\mu_J$ the average of J ;
- $\sigma_I^2$ the variance of I, $\sigma_J^2$ the variance of J, $cov_{IJ}$ the covariance of I and J ;
- $C_1 = (k_1L)^2$, $C_2 = (k_2L)^2$ two variables intended to stabilize the division when the denominator is very weak ;
- L = the dynamics of the pixels values, 255 for images coded on 8 bits ;
- $k_1 = 0,01$ and $k_2 = 0,03$ by default.

PSNR and SSIM are not suitable for SVD compression. Indeed, the typical value for PSNR is 30 dB and for SSIM, it is 0,93. However, a 30 dB compressed image with SVD is far from being a good quality image. The following images illustrate this perceptible distortion with a compressed image compared with the original image.

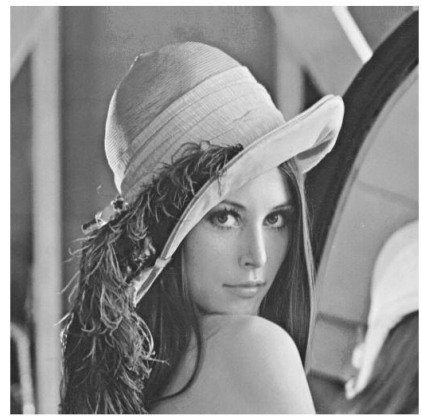

(a)

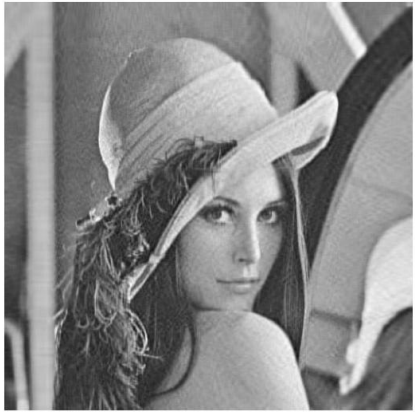

(b)

Fig. 1. (a) Original image, (b) Compressed image with PSNR = 30 dB, SSIM = 0,9352

In addition, PSNR is an indicator that does not consider the Human Visual System because it computes only the difference pixel-by-pixel of the two images. And in spite of SSIM is considered to reproduce the Human Visual System, its formula based on the averages and standard deviation shows that its authors were inspired by PSNR.

## 2. Proposed Metric

Seen the critics and discussions about the PSNR and SSIM performances, in order to evaluate the quality of the SVD compressed image, we propose the energy ratio corresponding to the ratio between the energy of the compressed image Ik and the original image I.

$$E = \frac{\|I_k\|}{\|I\|} = \frac{\sum_{i=1}^{k}\sigma_i^2}{\sum_{i=1}^{n}\sigma_i^2} \qquad (9)$$

The energy ‖I‖ coincides with the Hilbert-Schmidt norm.

We consider that if the energy E is restored to more than 99 %, the image will be considered good quality. In addition, as the energy ratio is between 0 and 1, it will be easy to compare it with SSIM.

## 3. Results

We compared the SSIM and E by varying the number of singular values retained for the test images « lena, liftingbody and mandrill » of size 512 × 512 pixels.

The following graph (Fig. 2) describes the variation of SSIM and E according to k.

We notice a small variation of the energy ratio E contrary to SSIM. In addition, E is always between 0,90 and 1.

Figure 3 shows that from k = 40, we already obtain images with 99,9 % restored energy.

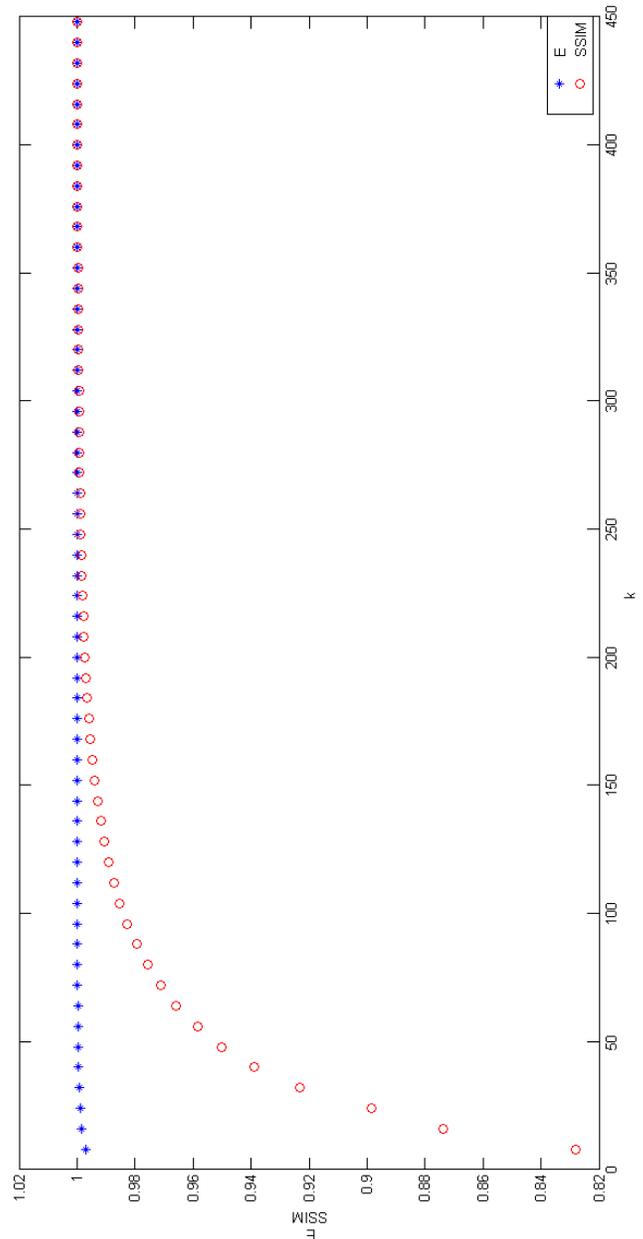

Fig. 2. The variation of SSIM and E according to k.

We notice also the presence of 3 different areas (Fig. 3, 4 and 5) corresponding to 3 different quality levels.

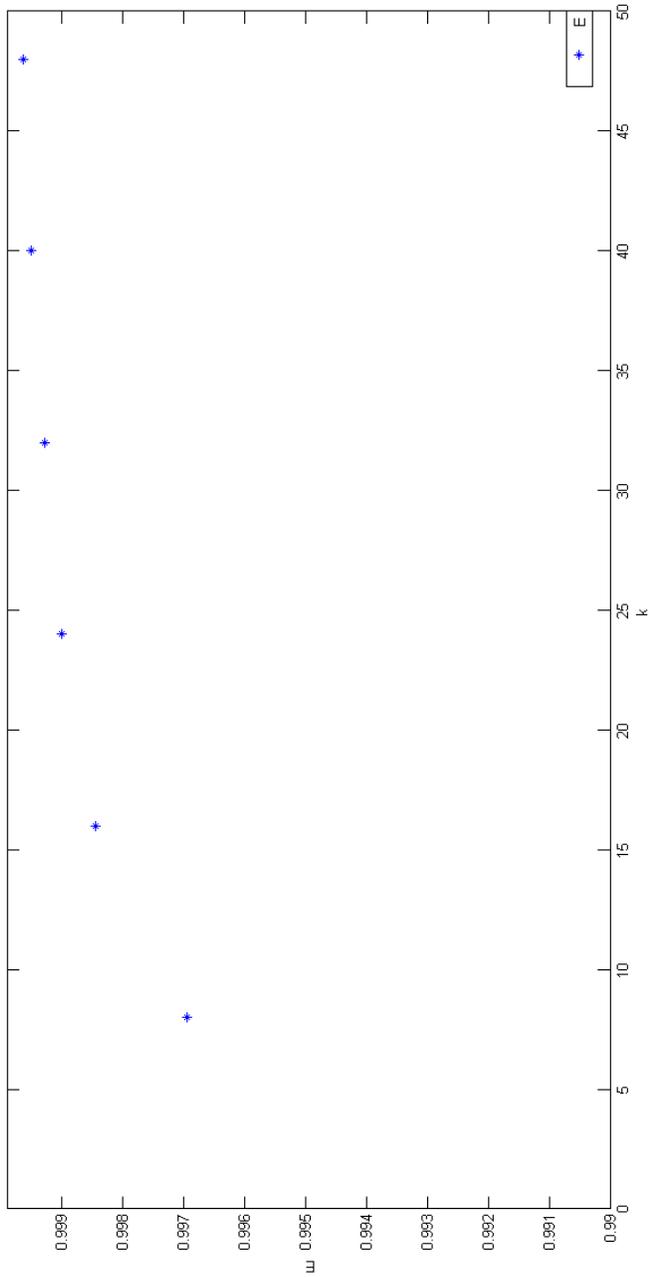

Fig. 3. E (between 0,99 and 0,999) according to k (AREA 1)

The first area corresponds to E between 0,99 (99%) and 0,999 (99.9%).

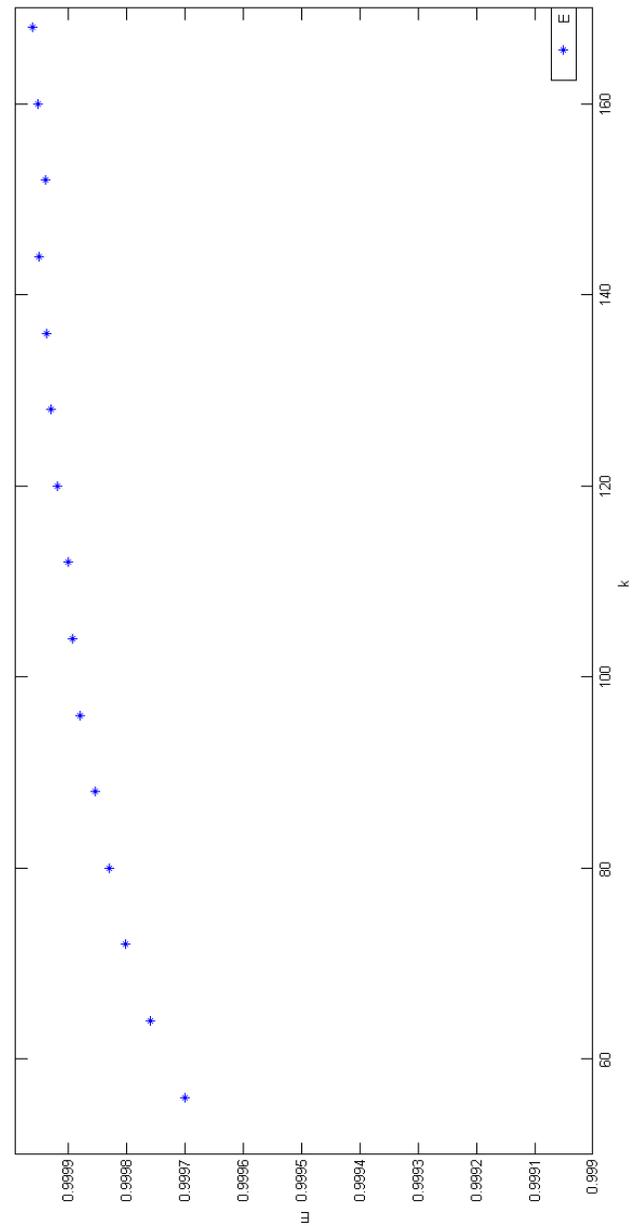

Fig. 4. E (between 0,999 and 0,9999) according to k (AREA 2)

The second area corresponds to E between 0,999 (99,9%) and 0,9999 (99,99%).

Table 1. Appreciation areas based on energy ratio

| k | 8 to 32 | 40 to 120 | 128 to 448 |
|---|---|---|---|
| PSNR | 27 dB to 34 dB | 35 dB to 42 dB | 43 dB to 98 dB |
| SSIM | 0,82 to 0,93 | 0,94 to 0,98 | 0,98 to 1 |
| E | 99,39 to 99,85 | 99,9 to 99,98 | 99,99 to 100 |
| Zone | 99 | 999 | 9999 |
| Appreciation | Poor quality | Good quality | Very good quality |

We obtain poor quality images for $8 \leq k \leq 32$ corresponding to « TWO NINE » ; good quality for $40 \leq k \leq 120$ corresponding to « THREE NINE » ; very good quality corresponding to $k \geq 128$ « FOUR NINE ».

The following figure shows the good quality of an image coded with E = 99,9 % corresponding to a PSNR of 35 dB.

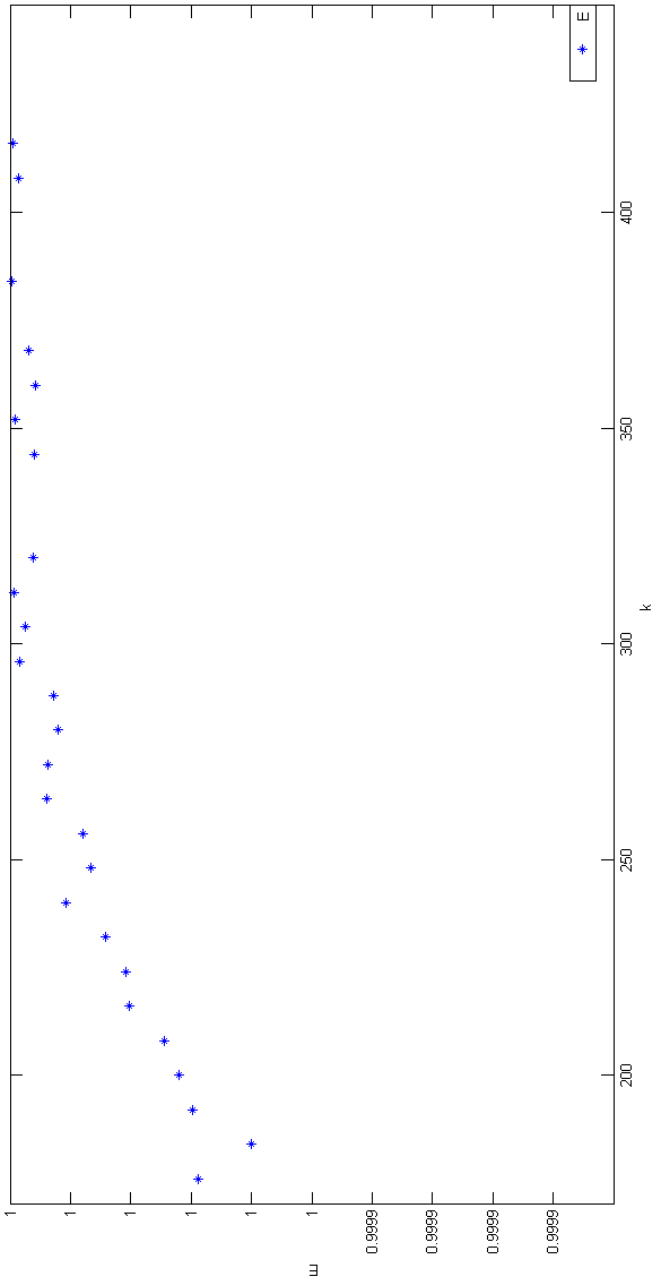

Fig. 5. E (between 0,9999 and 1) according to k (AREA 3)

The third area corresponds to E between 0,9999 (99,99%) and 1 (100%).

The following summary table describes these 3 areas with the average values of PSNR, SSIM and E obtained for all the test images.

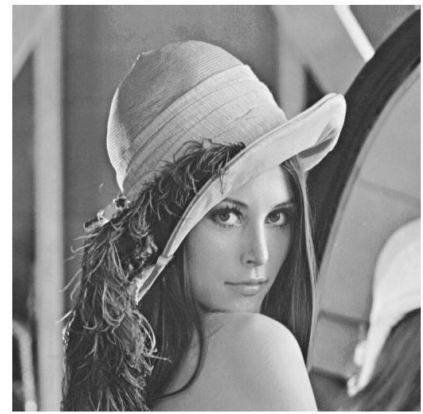

(a)

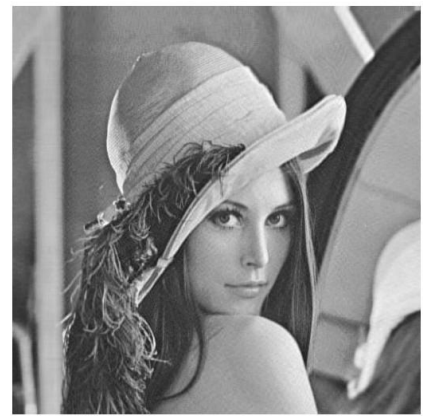

(b)

Fig. 6. (a) Original image, (b) Compressed image with PSNR = 35 dB, SSIM = 0,94

## 4. Conclusion and Perspective

Energy Radio can be used to complete the weaknesses of PSNR and SSIM that are only visual indicators. The second area coincides well with the threshold of appreciation of the PSNR which says that between 30 dB and 50 dB, the image is good quality. But the gap from 5 dB indicates that with the algorithm SVD, the typical value of the PSNR is 35 dB. As the difference between $E_{min}$ and $E_{max}$ is lower than 0,1; we can say that the metrics is very accurate with regard to SSIM and PSNR. The energy ratio is a simple indicator to use but require an important calculation time. It could be improved by using the parallel algorithm [8] proposed by Saira Banu. The next step will be to apply the energy ratio to the color images. We could also use it to evaluate the hybrid algorithms such as SVD and MPQ-BTC [9].

## References

[1] A. B. Watson, « Image Compression Using the Discrete Cosine Transform », Mathematica Journal, 1994, pp. 81-88.
[2] S. C. Meadows, « Color image compression using wavelet transform », Thesis in Electrical Engineering, 1997, pp. 1-86.
[3] J. Chen, « Image compression with SVD », ECS 289K Scientific Computation, 2000, pp. 13
[4] A. Abrahamsen and D. Richards, « Image Compression using Singular Value Decomposition », Linear algebra applications, 2001, pp. 1-14.
[5] S. Kahu, R. Rahate, « Image Compression using Singular Value Decomposition », International Journal of Advancements in Research & Technology, 2013, Volume 2, Issue 8, pp. 244-248.
[6] A. R. Sadek, « SVD Based Images Processing Applications : State of the Art, Contributions and Research Challenges », International Journal of Advanced Computer Science and Applications, 2012, Volume 3, No. 7, pp. 26-34.
[7] Z. Wang, A. C. Bovik, H. R. Sheikh and E. P. Simoncelli, « Image quality assessment :From error visibility to structural similarity », IEEE Transactions on Image Processing, 2004, Volume 13, no. 4, pp. 600-612.
[8] J. SairaBanu, B. Rajasekhara and R. Pandey, « Parallel Implementation of Singular Value Decomposition in Image Compression Using Open Mp and Sparse Matrix Representation », Indian Journal of Science and Technology, 2015, Volume 8 (13), pp. 1-10.
[9] N. K. El Abbadi and Al, « Image Compression based on SVD and MPQ-BTC », Journal of Computer Science, 2014, Volume 10, pp. 2095-2104.

**Author Profile:**

**Henri Bruno Razafindradina** was born in Fianarantsoa, Madagascar, on 1978. He received, respectively, his M.S degree and Ph.D in Computer Science and Information Engineering in 2005 and 2008. He served since 2010 as a professor at Higher Institute of Technology Diego Suarez, became an assistant lecturer in 2011. His current research interests include images compression, multimedia, computer vision, information hiding.

**Nicolas Raft Razafindrakoto** is a professor at the Higher Polytechnic School of Antananarivo. His current research interests include petri networks and computer science.

**Paul Auguste Randriamitantsoa** is a professor at the Higher Polytechnic School of Antananarivo. His current research interests include automatic and computer science.